\documentclass[conference]{IEEEtran}
\IEEEoverridecommandlockouts
\usepackage{cite}
\usepackage{subcaption}
\usepackage{amsmath,amssymb,amsfonts}
\usepackage{algorithmic}
\usepackage{graphicx}
\usepackage{textcomp}
\usepackage{breqn}
\usepackage{hyperref}

\usepackage{xcolor}
\def\BibTeX{{\rm B\kern-.05em{\sc i\kern-.025em b}\kern-.08em
    T\kern-.1667em\lower.7ex\hbox{E}\kern-.125emX}}
\begin{document}

\title{Systematic Generalization in Neural Networks-based Multivariate Time Series Forecasting Models \\
\thanks{\textsuperscript{*} Equal Contribution}
}

\author{\IEEEauthorblockN{Hritik Bansal*}
\IEEEauthorblockA{\textit{Electrical Engineering} \\
\textit{IIT Delhi}\\
New Delhi, India \\
hbansal10n@gmail.com}
\and
\IEEEauthorblockN{Gantavya Bhatt*}
\IEEEauthorblockA{\textit{Electrical and Computer Engineering} \\
\textit{University of Washington}\\
Seattle, USA \\
gbhatt2@u.washington.edu}
\and 
\IEEEauthorblockN{Pankaj Malhotra}
\IEEEauthorblockA{\textit{TCS Research} \\
New Delhi, India \\
malhotra.pankaj@tcs.com}
\and 
\IEEEauthorblockN{Prathosh A.P.}
\IEEEauthorblockA{\textit{Electrical Engineering} \\
\textit{IIT Delhi}\\
New Delhi, India \\
prathoshap@iitd.ac.in}
}

\maketitle

\begin{abstract}
Systematic generalization aims to evaluate reasoning about novel combinations from known components, an intrinsic property of human cognition. In this work, we study systematic generalization of NNs in forecasting future time series of dependent variables in a dynamical system, conditioned on past time series of dependent variables, and past and future control variables. We focus on systematic generalization wherein the NN-based forecasting model should perform well on previously unseen combinations or regimes of control variables after being trained on a limited set of the possible regimes. For NNs to depict such out-of-distribution generalization, they should be able to disentangle the various dependencies between control variables and dependent variables. 
We hypothesize that a modular NN architecture guided by the readily-available knowledge of independence of control variables as a potentially useful inductive bias to this end. 
Through extensive empirical evaluation on a toy dataset and a simulated electric motor dataset, we show that our proposed modular NN architecture serves as a simple yet highly effective inductive bias that enabling better forecasting of the dependent variables up to large horizons in contrast to standard NNs, and indeed capture the true dependency relations between the dependent and the control variables.
\end{abstract}


\section{Introduction}

The ability to learn the forward predictive model of state trajectories conditioned on past states, past actions, and future action trajectories using neural networks (NNs) is an important contribution of the deep learning revolution. The ability to learn such predictive models is leveraged in a wide variety of scenarios where the constituent objects behave in a highly complex fashion as a consequence of applied actions \cite{rw1}, \cite{rw15}, \cite{e7}. 
Additionally, there has been a lot of attention towards assessing the limitations of neural networks and understanding the extent to which they can generalize to novel scenarios that do not occur in the training distribution. Systematic generalization forms an essential part of such studies with its roots in the study of human cognition \cite{rw17}, \cite{rw18}. Being able to show systematicity implies that the models can compose the learned knowledge derived from training data in a meaningful way in new settings, e.g., the ability to understand ``walk right'' after training on ``right'', ``walk left'' and ``run right''. 

In this work, we focus on generalization ability of NN-based multivariate time series forecasting models. 
For multivariate time series resulting from various sensors measuring dependent variables or state and multiple controls of a dynamical system (such as an electric motor, engine, robotic arm with multiple degrees of freedom, etc.), the forecasting task is to predict the future time series of dependent variables given the past time series of control and dependent variables. In several real-world applications involving more than one control, only few pre-defined operating regimes or combinations of the control variables may be used to operate or control the dynamical systems.
The resulting data may thus not contain all possible ``combinations'' of the values that the control variables can take, posing challenges for learning forecasting models that can generalize well to previously unseen combinations. 
It is desirable to have forecasting models that can reason about all possible combinations of future controls (actions) despite being trained on a subset of them.

We analyze a fundamental question: what are the key architectural inductive biases that achieve better generalization than vanilla NN models? 
We intend to leverage a readily available knowledge of dynamical systems to our advantage: the control variables are independent of each other. This knowledge is in contrast to what the observational data might indicate where only certain combination of values of the control variables may be present in the available data due to pre-defined constraints on the operating regimes, thus depicting ``spurious correlations''. 
We aim to construct NN architectures that leverage this knowledge of independence of control variables to guide the learning of correct dependency relations from the data, and improve forecasting performance on previously unobserved operating regimes. 

Key contributions of this work can be summarized as follows:
\begin{enumerate}
    \item We highlight a practical challenge in data-driven modeling of dynamical systems, wherein, models tend to be biased to the operating regimes observed in the training data.
    \item We propose a novel modularized NN architecture that can leverage a readily available domain knowledge in form of independence of control variables, and in general, sparse dependencies across all variables in a dynamical system.
    \item Our proposed modularized NN architecture depicts i. better systematic generalization of the NN forecasting models using the proposed approach to unobserved operating regimes or control variable combinations, ii. better learned dependencies between control variables and dependent variables, and iii. better long horizon forecasts, in contrast on vanilla NN architectures without the proposed inductive biases. 
\end{enumerate}


The rest of the paper is organized as follows: we define the problem setup in Section \ref{pf}, followed by review of relevant literature in Section \ref{rw}. We provide details of various inductive biases in our modularized NN architecture in Section \ref{architecture}, followed by empirical analysis and observations in Section \ref{experiments}. We finally end with a conclusion and discussion in Section \ref{conc}.

\begin{figure*}[h]
  \centering
    \includegraphics[scale=0.50]{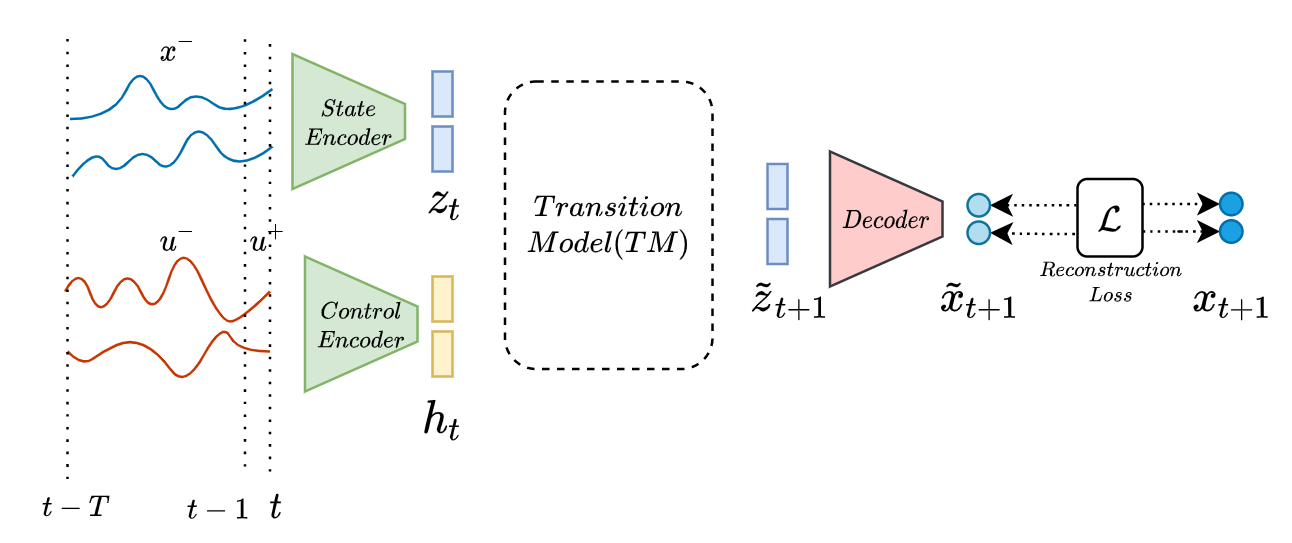}
  \caption{Illustration of various modules in the proposed architecture for 1-step ahead forecasting. Temporal segments of state and control variables are first encoded, then passed through a transition model, and then the future states are estimated using a linear decoder.}
  \label{fig:pipeline}
\end{figure*}

\section{Problem Formulation\label{pf}}

Consider a dataset of $N$-dimensional multivariate time series from a dynamical system, i.e., $\mathbf{x}_{t} = \{{x}_{1}^t, \ldots, {x}_{K}^t\} \in \mathcal{R}^{K}$, $\mathbf{u}_t = \{{u}_{1}^t, \ldots, {u}_{N-K}^t\} \in \mathcal{R}^{N-K}$ such that $t = \{1, \ldots, T\}$ where $x^t_{i}$ represents $i^{th}$ state variable, and $u^t_{j}$ represents $j^{th}$ control variable at time $t$, and $T$ is the length of time series.
The changes in the value of state (or dependent) variables, $\mathbf{x}$, are guided by the values of the control variables, $\mathbf{u}$, which in-turn are independent of each other.

Similar to \cite{rw1}, we consider a dynamical system with $\mathbf{x}_t$ as the state (value of dependent variables) of the dynamical system, and $\mathbf{u}_t$ as the control input. Our goal is to efficiently learn the future dynamics of this system, which robustly predicts over the long horizons even in the presence of sensory noise and delays in the control input. 
We consider $T$ previous inputs in addition to the current states: $x^{-} = \{\textbf{x}_{t-T}, \ldots ,\textbf{x}_{t-1}, \textbf{x}_{t}\}$, and similarly previous and current control inputs: $u^{-} = \{\textbf{u}_{t-T}, \ldots ,\textbf{u}_{t-1}\}$, $u^{+} = \{\textbf{u}_{t}\}$, to predict the future state $\mathbf{x}_{t+1}$. Including a temporal segment of previous control inputs also helps in the situation when the system has a delay in the control inputs. In this paper, we consider the dimension of state $\textbf{x}$ to be equal to that of the control $\textbf{u}$. 

The underlying dependency relations influence the way in which a single or multiple control variable(s) may influence the value of a single or multiple state variable(s). While training a NN model to predict future trajectories of the state variable, it is often a case that only some combinations of the control variables are observed. The observable combinations of control variables are treated as \textit{in-distribution} (IID) operating regime, while the novel (unseen) combinations are treated as \textit{out-of-distribution} (OOD) operating regime.

\section{Related Work\label{rw}}

\textbf{Systematic Generalization:} Systematic generalization has recently gained a lot of traction due to its ability to uncover neural networks' compositional capabilities. The work by \cite{rw4}, \cite{rw6} suggests that sequence-to-sequence models test poorly on the data points that require compositional skills and latch onto data-specific regularities. The work on visual question answering models by \cite{rw5} corroborates the negative generalization capabilities of generic models and highlights that systematicity can be improved by adding explicit structure to deep learning models. Similarly, \cite{rw7} study systematic generalization with Transformer-based models. Despite being a ubiquitous concept in the domain of natural language, we aim to extend it to the neural networks used for time-series modeling in this paper.

\textbf{Time Series Forecasting:} There have been various models such as autoregressive models and state-space models used for time series forecasting that require domain-specific rules to be known or require assumptions made regarding the property of the time series \cite{rw8}. With the advent of deep learning, various neural network architectures have been successfully able to forecast in highly non-linear time series spaces \cite{rw9}, \cite{rw10}, \cite{rw12}, \cite{e4}. In this work, we ask a finer question: what key architectural choices enable systematic generalization and capture some level of underlying interactions between the states and the controls? 

\textbf{Representation Learning:} The ability to learn high-level representations from the data with neural networks in an end-to-end fashion is also a very popular idea \cite{rw11}. Recent work proposes different methods to learn useful representations of the time series data in an unsupervised way that can be leveraged to perform well on downstream tasks such as classification \cite{rw2}, \cite{rw3}. These methods choose to have an additional contrastive loss term, which has shown to be useful for representation learning across various applications \cite{rw13}, \cite{rw14}, \cite{e7}. However, in this paper, we purely evaluate the models' systematic generalization capabilities with different architectural inductive biases and leave the deployment of additional loss terms to improve systematicity as future work.

\textbf{Action-Conditioned State Trajectory Prediction:} The problem of learning dynamics of a system to predict state trajectories conditioned on the applied control input (action) is a primary focus of robotics and reinforcement learning \cite{rw15}. \cite{rw1} presented an approach to learning the distribution over future state trajectories conditioned on past states, a past action, and future actions using variational autoencoder. However, our formulation does not consider stochasticity in environment transitions or observations, hence limited to deterministic dynamics world models. The models need to reason computationally in terms of states, controls, and their dependencies for them to predict future dynamics of multiple states well \cite{e7}. Our work analyzes the effect of strong inductive biases in the latent transition models to predict future state trajectories.

\begin{figure*}[h]
  \centering
    \includegraphics[scale = 0.60]{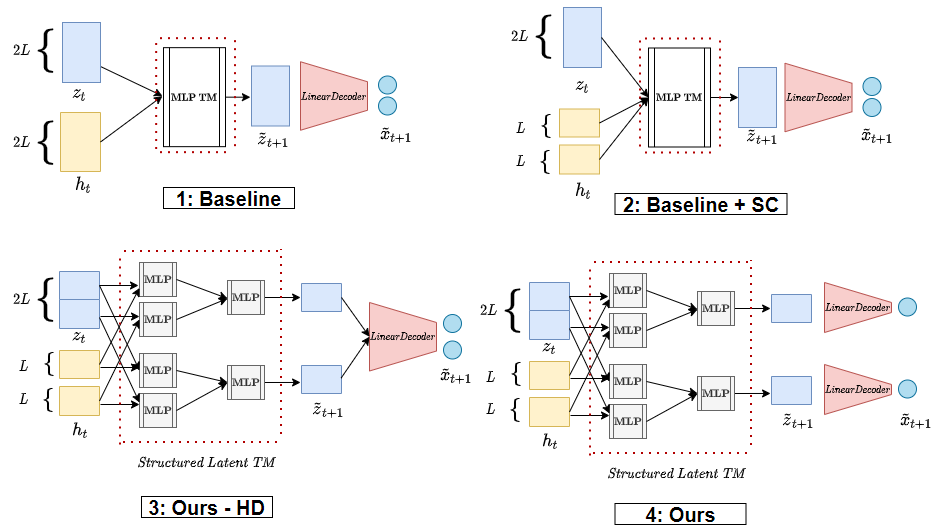}
  \caption{Variants of Transition Model (TM) and Decoder modules considered. In variants 1 and 2, we use MLP as the TM, while in variants 3 and 4 we use a structured latent model which depicts better systematic generalization. }
  \label{fig:TM}
\end{figure*}


\section{Approach}

\label{architecture}

We consider a modularized neural network architecture with three core modules: state and control encoders, a transition model, and a linear decoder (Figure \ref{fig:pipeline}). 
We next motivate various possible inductive biases or design choices for each of the modules assuming modeling of a dynamical system with two state variables and two control variables.

\subsection{State and Control Encoder}
We encode the state and control time series via separate encoders. Each encoder comprises a 1D causal convolutional neural network (causal-CNN). Causal convolutions map a sequence to a sequence of the same length such that the $i^{th}$ output sequence is calculated using the values up till $i^{th}$ element of the input \cite{rw2}. Each dimension of the state variable contributes to one channel of the input to the CNN i.e., if $x$ is two-dimensional, the CNN would have two input channels. However, we consider two variants of encoding possibilities for the control variables. First, \textit{\textbf{common control}} encoder where each dimension of the control variable contributes to one channel of the input to the CNN, identical to the state encoder. Second, \textit{\textbf{separate control (SC)}} encoder where we have a separate causal CNN with one output channel for each control input time series. Separate control encoders do not share weights. In the later choice of encoding, we have used the fact that control inputs are exogenous variables and deserve the independent treatment of their own. We posit that having a common CNN encoder for the control variables learns spurious correlations between the controls. The number of output channels for the state encoder and common control encoder is kept to be equal to the dimension of the control variable.


\subsection{Transition Model}
After passing the state and control temporal segment through their respective encoders, we get their representations $\mathbf{z}_t$ and $\mathbf{h}_t$, respectively. Let the dimension of these representations for each state and control variable be L. Thus, a scenario of 2 state variables and 2 control variables would yield $\mathbf{z}_t$ and $\mathbf{h}_t$ to be of 2L dimension each. Consider the architectures 1-4 in Figure \ref{fig:TM}. In Architecture 1, we have a common control encoder, while in the others, we have a separate control encoder. 
We indicate this by displaying $\mathbf{h}_t$ for each input separately. We implement a transition model which takes these  $\mathbf{z}_t$ and $\mathbf{h}_t$, and predicts the representation for future state, $\mathbf{\tilde{z}}_{t+1}$. To test different inductive biases, we consider two major variants of the transition model. 

\subsubsection{MLP Transition Model}
The MLP transition model concatenates the representations $\mathbf{z}_t$ and $\mathbf{h}_t$. The concatenation operation is represented as two incoming arrows in Figure \ref{fig:TM}. Following the concatenation, the transition model passes them through a 2-layer MLP, producing a vector $\mathbf{\tilde{z}}_{t+1}$ of dimension 2L. In Figure \ref{fig:TM}, this transition model is presented as the \textbf{Baseline} model as it lacks any structural inductive bias. \textbf{Baseline + SC} is identical to baseline, except it contains control embeddings coming from separate causal CNN encoders.  
\subsubsection{Structured Latent Space Transition Model}
In this case, we intend to have a useful inductive bias in form of factored representation space that is less prone to learning spurious correlations as it structurally resembles the kind of rules it is supposed to learn from the data. A good transition model is necessary to forecast well and improve performance on other downstream tasks. We empirically verify the usefulness of such inductive bias in the Section \ref{experiments}.

The structured latent space allows for a systematic exchange of information between the state and control inputs, as shown in the Figure \ref{fig:TM}. Let us assume a scenario where we have $n$ state and $n$ control variables. Hence, the state encoder and control encoder output $n$ state representations, $z_{t}(i)$, and $n$ control representations, $h_{t}(i)$, respectively, such that the dimension of $h_{t}(i)$ and $z_{t}(i)$ is L.

Firstly, we get $E_t^j = \{e_{t}^{ij}, i \in \{1, 2, \ldots, n\}\}$, for every $n$ state and control variable as:  
\begin{equation}
    e_{t}^{ij} = f_{1}^{ij}([z_t(i); \ h_t(i)]) \hspace{0.4cm} \forall j \in {1,2, \ldots n},
\end{equation}
where $;$ represents the concatenation operator, $f_{1}^{ij}$ represents a 1-layer MLP which maps the input of dimension 2L to a output of dimension L. Here, $e_t^{ij}$ captures the dependencies between the state and control representations to forecast the trajectory of state $x_{t+1}^j$. We collate all such dependency relations, and pass them through another 1-layer MLP, $f_{2}^{j}$ to estimate the representation of the future states:
\begin{equation}
    \tilde{z}_{t+1}(j) = f_{2}^j([e_{t}^{1j}; e_{t}^{2j}; \ldots; e_{t}^{nj}]) \hspace{0.4cm} \forall j \in {1,2, \ldots n}.
\end{equation}
Thus, $f_{2}^{ij}$ takes the input of $nL$ dimension to output dimension of $L$. Figure \ref{fig:TM} presents this architecture as \textbf{Ours}. All of $f_{1}^{ij}$ and $f_{2}^{ij}$ correspond to a different neural network.

\subsection{Decoder}
After we get $\mathbf{\tilde{z}}_{t+1} = [\tilde{z}_{t+1}^{1}; \tilde{z}_{t+1}^{2}; \ldots;\tilde{z}_{t+1}^{n}]$ from the transition model, we pass it through a decoder, to obtain the forecast or prediction for the next states, $\mathbf{\tilde{x}_{t+1}}$. Based on the way we decode, we construct two possibilities. Firstly, we have \textbf{\textit{common decoding}}, where we directly pass $nL$ dimensional $\mathbf{\tilde{z}}_{t+1}$ through decoder to generate a $n$-dimensional $\mathbf{\tilde{x}}_{t+1}$. Here, reconstruction loss is calculated between the forecasted states ($\mathbf{\tilde{x}}_{t+1}$) and the ground truth trajectory ($\mathbf{{x}}_{t+1}$). This loss gets back-propagated through the complete pipeline, and trains it end-to-end. This decoding scheme is applicable for both the MLP transition model and the structured latent transition model. On the other hand, we define a \textbf{hard decoder (HD)}, where we pass the representation for each forecasted state variable $\tilde{z}_{t+1}(j)$ from structured latent transition model to separate decoders. Across all the settings, the decoder is implemented using one fully connected network\footnote{In case when we have a mismatch in the number of states and control input, we will use a 1D CNN decoder, with the number of input channels being the number of control variables and the number of output channels being the number of state variables. However in this case we only have a common decoder. }. This model is presented as \textbf{Ours - HD} in Figure \ref{fig:TM}.

\subsection{Multi-step ahead forecasting}
In our framework, we train the models for multi-step ahead forecasting. In the previous section and Figure \ref{fig:pipeline}, we have described the pipeline for one-step forecasting. For multi-step forecasting, the future state representations $\mathbf{\tilde{z}}_{t+1} \ldots \mathbf{\tilde{z}}_{t+M}$ are obtained by starting with $\mathbf{z}_t$ and
$\mathbf{h}_t$ as initial inputs to the transition model to obtain $\mathbf{\tilde{z}}_{t+1}$, and iteratively using $\mathbf{\tilde{z}}_{t+i}$ along with the corresponding  $\mathbf{h}_{t+i}$ (estimated using $\mathbf{u}_{t+i-T} \ldots \mathbf{u}_{t+i}$) as inputs to the transition model ($i=1\ldots M$). 
The obtained $\mathbf{\tilde{z}}_{t+1} \ldots \mathbf{\tilde{z}}_{t+M}$ are mapped to the corresponding output estimates $\mathbf{\tilde{x}}_{t+1} \ldots \mathbf{\tilde{x}}_{t+M}$ via the decoder. 
The multi-step loss with horizon $M$ is given by
$\mathcal{L} = \frac{1}{M} \sum_{k=1}^{M} \| \mathbf{x}_{t+k} - \mathbf{\tilde{x}}_{t+k} \|_{2} ^ {2}$.

\section{Experiments}
\label{experiments}

Our goal of this experimental section is to study the effect of various inductive biases on their ability to 1) learn a good transition model in the latent space that helps in 2) accurate forecasting up to large horizons, and 3) generalizes systematically to an unseen combination of future actions (control variables).

\subsection{Experimental Setup}
To compute the loss function, we train our network for 5-step ahead forecasting task. All the models are trained for ten different random seed initializations with ADAM optimizer \cite{ts1}. We consider four versions of the architecture, based on the type of control encoder- common control v/s separate control, type of transition model- MLP v/s structured latent transition model, and decoder choice common v/s hard decoding. We write separate control as SC and hard decoding (or separate decoding) as HD. 
Other training settings and the choice of hyperparameters are available at our codebase\footnote{\href{https://github.com/Hritikbansal/MultivariateTimeSeries_Generalization}{https://github.com/Hritikbansal/MultivariateTimeSeries\_Generalization}}.

\subsection{Synthetic Data}
\label{synthetic}

We obtain multivariate time series using Non-Linear Autoregressive Moving Average (NARMA) \cite{e2} generators from TimeSynth Library\cite{e1}. For our purpose, we use the following modified dynamical equation of NARMA signals:
\begin{equation}
\label{narma eqn}
    x_{i,t+1} = a_{i}x_{i,t} + b_{i}x_{i,t}\sum_{k = t-m}^{t}{x_{i,k}} + \sum_{j = 1}^{n}{c_{i,j}}{u_{j,t}}{u_{j,t-m}} + d_{i}
\end{equation}

where $x_i$ is the state (dependent or endogenous) variable in the $i^{th}$ dimension, $u_{j}$ is the control (exogenous) variable in the $j^{th}$ dimension, n equals the dimension of the state and control variable in our framework ($1 \leq i \leq n, 1 \leq j \leq n$), m is the order of non-linear interactions, t is the current timestamp, $c_{i,j}$ controls the interaction between the control $u_j$ and the state $x_i$ (\ref{scenarios}), and $a_i$, $b_i$, $d_i$ are parameters specific to state $x_i$. In our experiments, we fix n = 2 which makes it easy to manipulate different state and control variable relations, and have fixed m = 10 for our experiments.  

We deliberately introduce spurious correlations between the two (n = 2) controls using:
\begin{equation}
\label{narma control eqn}
    u_{2,t} = \alpha u_{1,t} + ( 1 - \alpha )k  \hspace{1cm}\forall t
\end{equation}
where $k \sim U(0,1)$, $u_{1,t} \sim U(0,1)$, and $\alpha \sim U(p,q)$. Here $\alpha$ acts as a correlation factor between the two control variables, tweaking which allows us to test generalization to unseen future control combinations (\ref{iid ood narma}).

\subsubsection{NARMA Scenarios}
\label{scenarios}
\begin{figure}[h]
  \centering
    \includegraphics[scale=0.60]{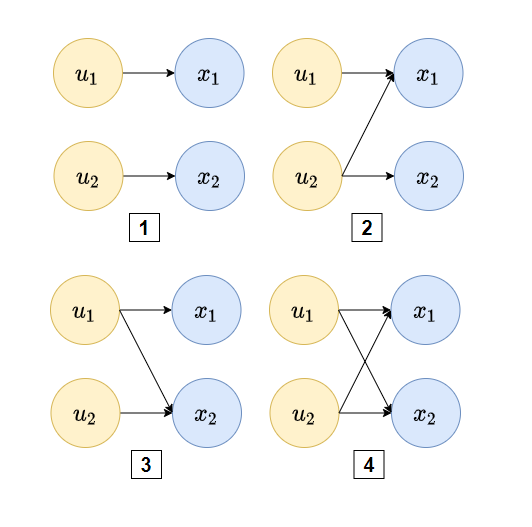}
  \caption{We consider four scenarios in our synthetic dataset. Out of the four scenarios, it is most difficult to learn correct dependencies in the first one, which we show using out of distribution testing examples.}
  \label{fig:scenario}
\end{figure}

From Equation \ref{narma eqn}, it is clear that $c_{i,j}$ determines the causal relations between the states and the controls across dimensions e.g., $c_{1,2} = 0$ and $c_{2,1} \neq 0$ implies that $u_{2}$ does not effect $x_{1}$ but $u_{1}$ effects $x_{2}$. Following the same principle, we create four NARMA scenarios in our work (Figure \ref{fig:scenario}). These scenarios allow us to assess which dependency relations are better modelled by the learned latent transition model. 

Scenario 2 differs from Scenario 3 by the strength of causal relation between the controls and states governed by parameter $c_{i,j}$. We keep $(c_{1,1}, c_{1,2}, c_{2,1}, c_{2,2}) = (1.5, 0, 0.5, 0.55)$ for Scenario 2, and $(c_{1,1}, c_{1,2}, c_{2,1}, c_{2,2}) = (1.5, 0.15, 0, 0.55)$ for Scenario 3. Among all the four scenarios considered, Scenario 1 is the most challenging one as the models cannot predict the future trajectory of its states without learning the underlying interactions perfectly. It is important for the models to capture the true underlying dependencies to forecast well up to large horizons. 

\subsubsection{Systematic Generalization Setup}
\label{iid ood narma}
\begin{figure}[h!]
  \centering
    \includegraphics[scale=0.3]{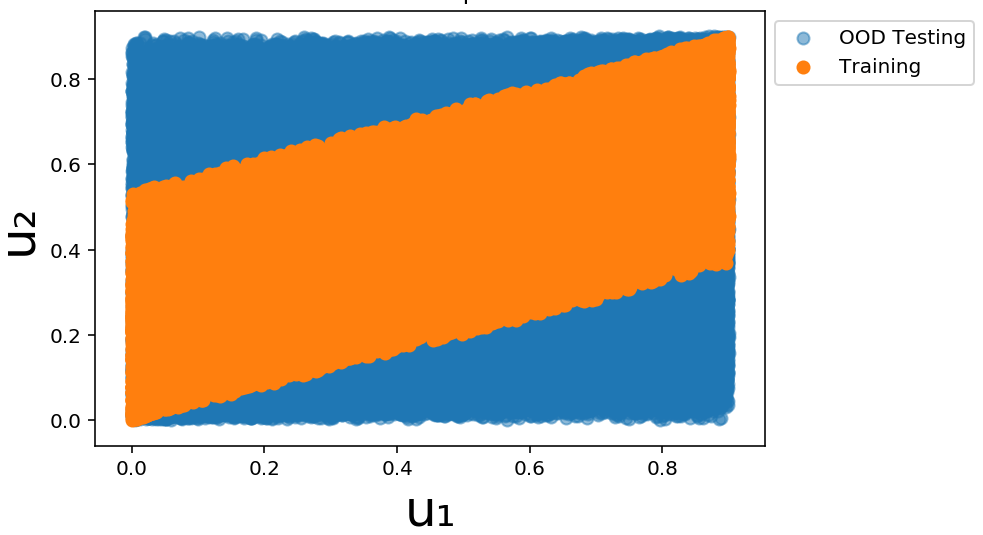}
  \caption{Distribution of control inputs. During the training, we expose the models with the control inputs in the orange region. At the test time, we evaluate for the inputs in the blue region (the upper and lower triangles) }
  \label{fig:control distribution}
\end{figure}

In addition to performing well on control combinations, $(u_{1}, u_{2})$, that have been observed in the training phase, a robust model should be able to reason about unseen combinations of future controls after being trained on all of them i.e., observing all individual values of $u_{1}$ and $u_{2}$. Figure \ref{fig:control distribution} shows the joint distribution of control inputs $(u_{1}, u_{2})$.

We test systematic generalization by restricting the combination of controls seen during training by choosing $\alpha \sim U(0.4, 0.7)$ in Equation \ref{narma control eqn} (Orange region in Figure \ref{fig:control distribution}). For out-of-distribution (OOD) testing, we generate controls by choosing $\alpha \sim U(0, 1)$. We ensure that the future controls lie outside the region seen by the models during the training (Blue region in Figure \ref{fig:control distribution}). All control inputs follow the identical setup across four scenarios described in \ref{scenarios}. 

\subsubsection{Results}
\label{narma results}

We set up quantitative experiments for evaluating the quality of the learned transition model (\ref{iid vs ood}), and the quality of state-control interactions captured by the models (\ref{noise}).

We train our models on 8k temporal segments over states and controls and test them on 2k segments. We fix the length of a segment T = 11 and train the models for 5 steps-ahead forecasting. Thus, five reconstruction losses are added to give the total loss, which gets back-propagated to make a single update to the whole pipeline. Multi-step prediction during training helps avoid learning sub-optimal transition models that could perform well on a single time step but may perform worse on larger horizons \cite{e5}.
We use Mean-Squared Error (MSE) to evaluate the quality of forecasts across different NARMA scenarios. We average report MSE across ten different random seeds for all the models in consideration.
\begin{table*}
\centering
    \resizebox{1\textwidth}{!}{%
\begin{tabular}{c|llll|llll|llll|}
          & \multicolumn{4}{c|}{MSE IID ($\times 10^{-5}$)}                   & \multicolumn{4}{c|}{MSE OOD ($\times 10^{-5}$)}                 & \multicolumn{4}{c|}{(OOD-IID)/IID}    \\
          \hline
NARMA Scenario & Baseline       & Baseline+SC    & Ours-HD & Ours & Baseline       & Baseline+SC    & Ours-HD & Ours & Baseline     & Baseline+SC & Ours-HD & Ours  \\
\hline
 1   & 0.89 & 0.71 & \textbf{0.69} & 0.80    & 27 & 3.18 & 2.75 & \textbf{2.74}    & 29.24 & 3.52 & 3      & \textbf{2.43}        \\
 2   & 1.30 & \textbf{1.02} & 1.08 & 1.33    & 31 & \textbf{3.82} & 4.67 & 5.29    & 22.71 & \textbf{2.73} & 3.34    & 2.97        \\
 3   & 1.46 & 1.06 & \textbf{0.94} & 1.00    & 23 & 3.53 & \textbf{3.08} & 3.50    & 14.86 & 2.35 & \textbf{2.29}    & 2.5         \\
 4   & 5.53 & 1.31 & \textbf{1.24} & 1.42    & 160 & \textbf{4.45} & 5.12 & 5.10    & 27.23 & \textbf{2.38} & 3.13    & 2.58        \\
\hline
\end{tabular}}
\caption{Results on a synthetic dataset. Note that models with structured latent TM  outperform the ones without MLP TM.}
\label{narma results}
\end{table*}
\begin{figure*} 
    \centering
  \subfloat[MSE of $x_1$ on perturbing $u_2$ with additive gaussian noise \\ of given standard deviation. \label{noise:1a}]{%
       \includegraphics[scale=0.5]{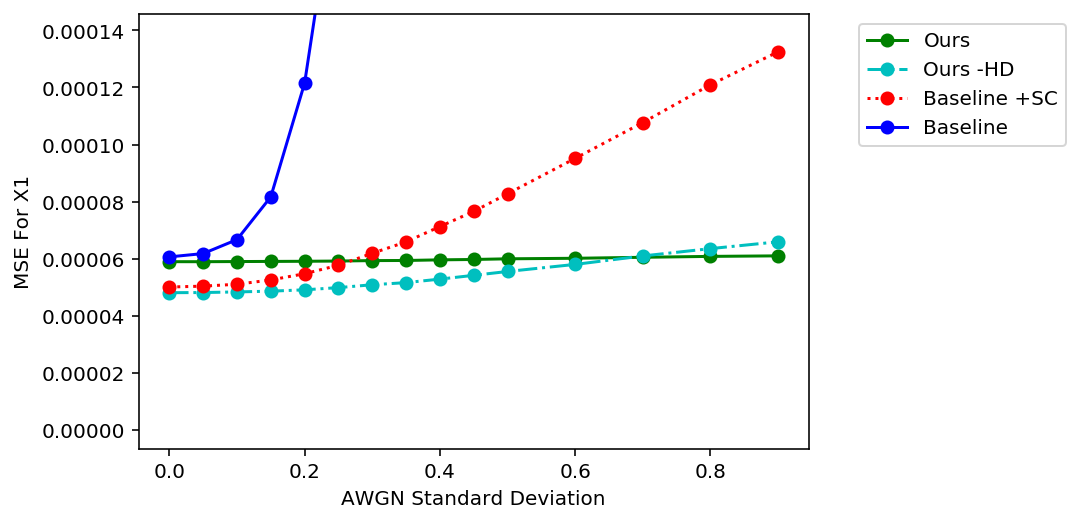}}
  \subfloat[MSE of $x_2$ on perturbing $u_1$ with additive gaussian noise \\ of given standard deviation.\label{noise:1b}]{%
        \includegraphics[scale=0.5]{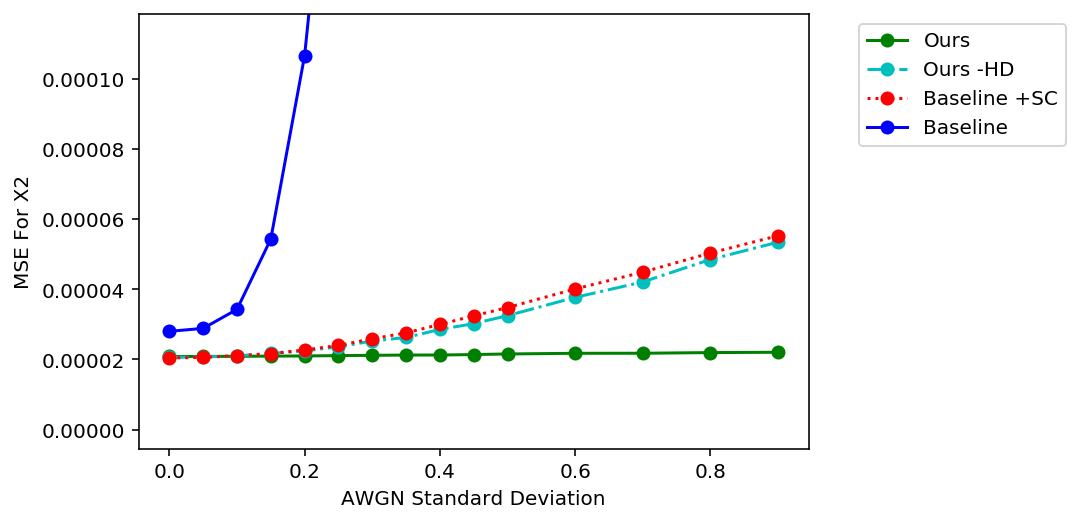}}
    \\
  \subfloat[MSE of $x_2$ on perturbing $u_1$ with additive gaussian noise \\ of given standard deviation.\label{noise:1c}]{%
        \includegraphics[scale=0.5]{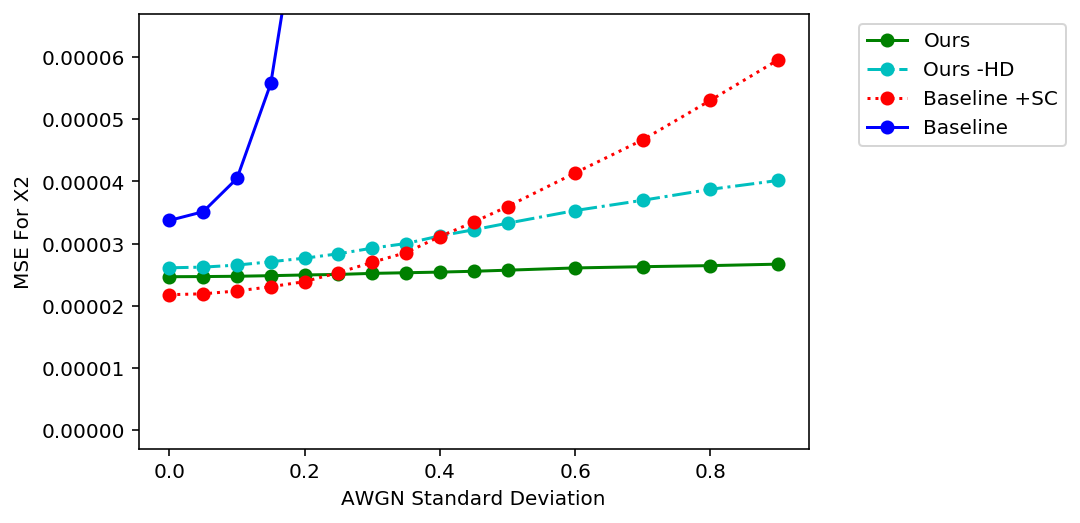}}
  \subfloat[MSE of $x_1$ on perturbing $u_2$ with additive gaussian noise \\ of given standard deviation.\label{noise:1d}]{%
        \includegraphics[scale=0.5]{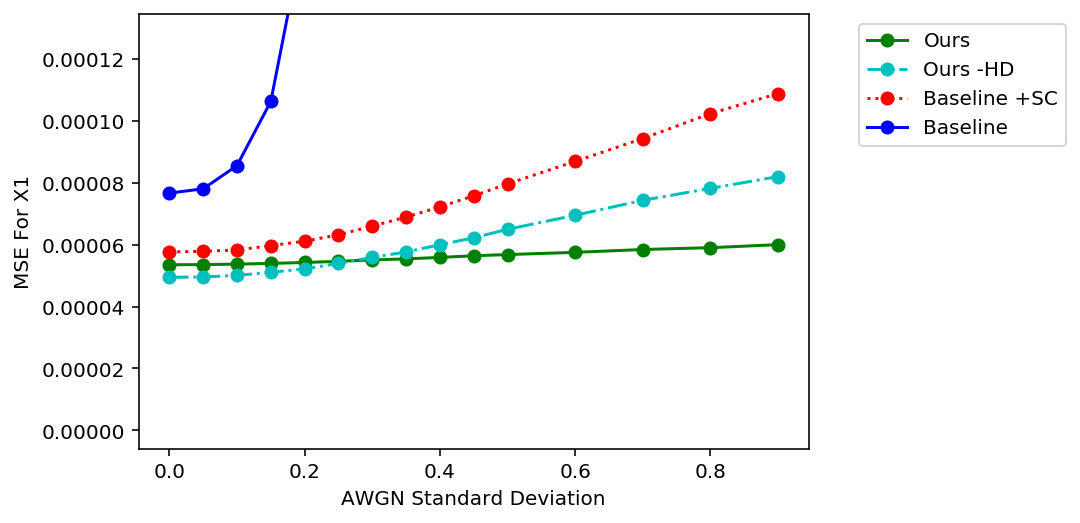}}
  \caption{Variation of MSE of the state variable, when adding white Gaussian noise on control input, which mathematically doesn't affect the state variable under test. We see performance of our model does \textit{not} change even under large perturbations. }
  \label{fig:noise graphs} 
\end{figure*}

\begin{table*}
\centering
    \resizebox{1\textwidth}{!}{%
\begin{tabular}{lll|lll|lll|lll|}
\multicolumn{3}{c|}{Baseline}         & \multicolumn{3}{c|}{Baseline+SC}        & \multicolumn{3}{c|}{Ours-HD}     & \multicolumn{3}{c|}{Ours}  \\
\hline
IID    & OOD  & (OOD-IID)/IID & IID    & OOD    & (OOD-IID)/IID & IID    & OOD    & (OOD-IID)/IID & IID    & OOD   & (OOD-IID)/IID  \\
\hline
0.014 & 0.19  & 12.89       & 0.011 & \textbf{0.015} & 0.36         & \textbf{0.010} & \textbf{0.015} & 0.44        & 0.011 & \textbf{0.015} & \textbf{0.32} \\        
\hline\\
\end{tabular}}
\caption{Results on electric motor dataset. Electric motor comes under the 4th scenario of our synthetic dataset.
}
\label{PMSM results}
\end{table*}

\paragraph{Forecasting -- IID vs OOD Performance}
\label{iid vs ood}

We provide results for the proposed model and the baseline in Table \ref{narma results}. The MSE values are reported for single-step forecasts during testing. Each row corresponds to a scenario from \ref{scenarios}. We divide the results into three attributes: forecasting MSE when the future controls are sampled from the same joint distribution as observed during the training phase (MSE IID), forecasting MSE when future controls lie outside observed joint distribution of future controls (MSE OOD), and relative error between MSE IID and MSE OOD.

From Table \ref{narma results}, it can be concluded that the baseline, the model without any factored representations in the latent transition model and without separate treatment to control encoders, performs worse (10x) on OOD testing across all scenarios. For the first three scenarios, this simplest architecture performed as well as the other architectures on the MSE IID metric. This confirms our evaluation procedure's robustness, as the model without appropriate inductive bias performs worse on MSE OOD and relative error metric. The MSE values are towards a higher end for all the models in scenario four as it possesses highly non-linear complex dynamics due to its dependency relations.  

The results also highlight the usefulness of the proposed factored latent transition model (\textbf{Ours}). In scenarios 1-3, the models need to prune the non-existent dependencies and avoid capturing false interactions for good forecasting performance. In 2 of these 3 scenarios, the models with structural transition model (\textbf{Ours}, \textbf{Ours - HD}) gave the least relative error. Additionally, it is interesting to see that the separate treatment of controls also improves the performance in the absence of a structural transition model. This indicates that having separate encoders for independent control variables guides the learning towards a better transition model.  

\paragraph{Intervention Testing} 
\label{noise}
By now, we have established that useful inductive biases help to generalize to unseen combinations through quantitative forecasting errors. However, it is not clear if the models are learning the correct relations between the states and the controls. To answer this question, we subject our models to intervention testing. From the theory of causality, \cite{e6}, if a variable X is not a cause of variable Y, then external intervention on X should not affect the values of Y, i.e., changing X should not change Y. As we already know the true causal relations in our NARMA scenarios (Figure \ref{fig:scenario}), we leverage intervention testing to assess what is being learned by the models.

In scenario 1, changing the value of future control $u_{2}^{+}$, using additive white gaussian noise (AWGN), should not change the prediction MSE for $x_{1}$, and similarly, wiggling $u_{1}^{+}$ should not change the MSE for $x_{2}$. An identical argument holds for scenario 2-3.

We present our results for four intervention experiments in Figure \ref{fig:noise graphs}. MSE values have been reported for 5-step forecasts averaged across models trained with ten different random seeds. From the graphs \ref{noise:1a} - \ref{noise:1d}, we can conclude that our approach with factored representations along with separate decoders (\textbf{Ours}) is most likely to have captured the underlying relations perfectly. Having separate decoders guides learning such that each $\tilde{z}_{t+1}(i)$ specializes in that state's dynamics, hence inducing disentangled representations. The MSE for the baseline architecture blows up after we slowly increase the standard deviation of the additive noise, indicating its poor ability to apprehend state-control interactions. For the model with separate control encoders and without factored representations (\textbf{Baseline+SC}), the MSE does not change much initially but increases gradually as the strength of intervention increases.
In contrast, the error increases less gradually for the model with structural inductive bias (\textbf{Ours-HD}). The best performing models have a separable latent structure that is well matched to the data's nature. Our observations may also help explain why graph neural networks (GNNs), which treat each object as a separate node, are well suited for learning good object-centric representations \cite{e7}, learning dynamic interactions \cite{e5} and discovering symbolic expressions \cite{e8}.

\subsection{Simulation: Electric Motor}
\label{simulation}

Modeling electric motors is a challenging problem because of its very complicated dynamics. Prior work has been done on using a purely data-driven approach to model electric motors without making any assumptions on its internal behavior \cite{e4}. We intend to evaluate our models on their ability to capture complex dynamics that are part of everyday life and analyze how well they can generalize to unseen combinations of controls in this context. Hence, we test our framework to a dataset (80k data points) generated using a three-phase permanent magnet synchronous motor (PMSM) model available at GEM toolbox \cite{e3}.

\subsubsection{Setup}
As described in \cite{e3}, PMSM consists of three phases, where each voltage has an associated phase voltage and a phase current. A series of transformations convert these three-phase voltages and currents into d-q coordinates fixed to the rotor. Hence, our dataset consists of currents $i_d, i_q$, voltages $u_d, u_q$ and rotor speed $\omega_r$. As we are using segment-based modeling, we make sure that rotor speed, $\omega_r$ remains constant across that segment of length T. This leaves us with $u_d, u_q$ acting as control variables and $i_d, i_q$ acting as state variables. Other physical quantities associated with motor dynamics such as flux, inductance, and resistance are fixed. 

\begin{figure}[h!]
  \centering
    \includegraphics[scale=0.30]{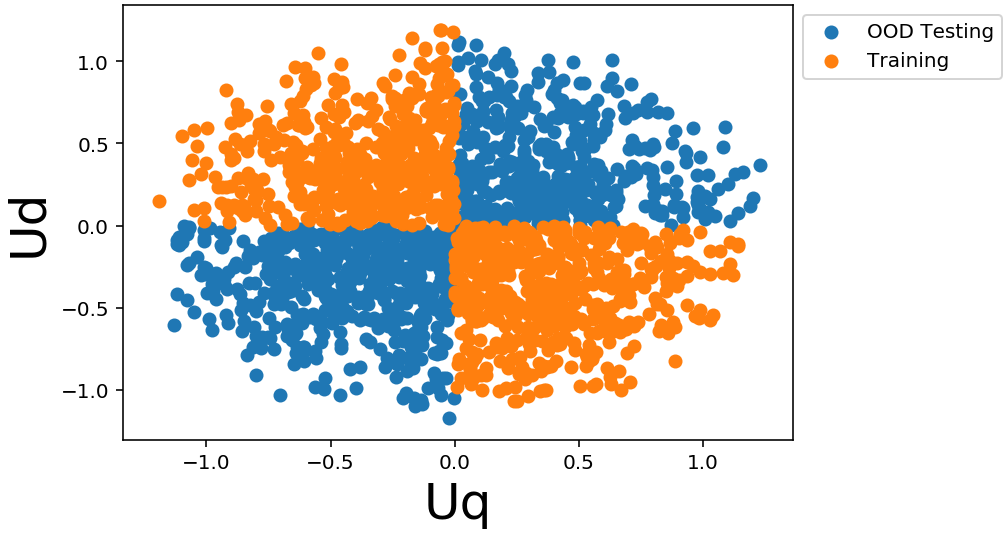}
  \caption{Distribution of control inputs for electric motor dataset.}
  \label{fig:pmsm distribution}
\end{figure}

Figure \ref{fig:pmsm distribution} shows the joint distribution of the controls ($u_d$, $u_q$) for PMSM dataset. To test a systematic generalization of our models, we make sure that they are only exposed to the first and third quadrant of the joint distribution of the control variables (Marked with Blue in Figure \ref{fig:pmsm distribution}). We make sure that the controls are sampled from the second and fourth quadrant only during out-of-distribution testing, enabling generalization testing to the unseen combination of ($u_d$, $u_q$).

\begin{figure}[h!]
  \centering
    \includegraphics[scale=0.45]{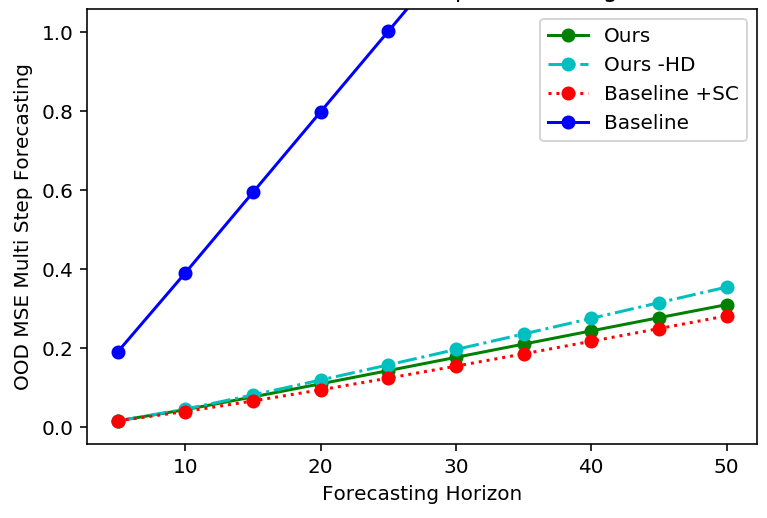}
  \caption{Variation of OOD loss with forecasting horizons.}
  \label{fig:pmsm multistep}
\end{figure}

\begin{figure} 
    \centering
  \subfloat[\label{ts_a}]{%
       \includegraphics[scale=0.45]{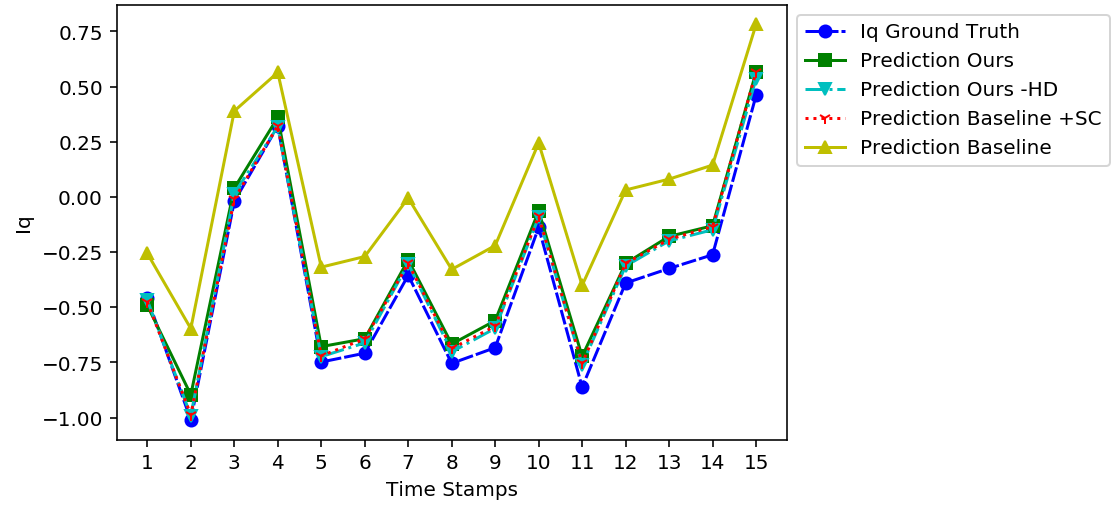}}
    \\
  \subfloat[\label{ts_b}]{%
        \includegraphics[scale=0.45]{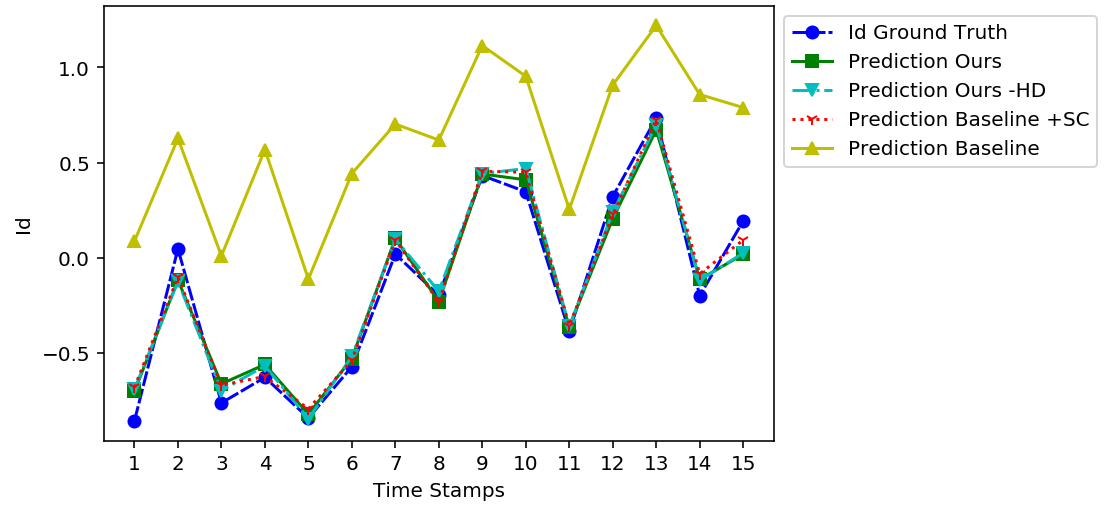}}
  \caption{Forecasting of $i_q$ and $i_d$ variable, when subjected to OOD control inputs, over a 15 time-stamp forecasting horizon.}
  \label{fig: timeseries} 
\end{figure}

\subsubsection{Results}
We train our models on the simulated PMSM system (4k temporal segments; T = 10) using 5-step forecasts over ten different random seed initializations. Testing is performed over 1k segments for this experiment. We evaluate if the latent transition models can capture real-world physical systems' complex internal dynamics and forecast up to large horizons while generalizing systematically.

\paragraph{Forecasting -- IID vs. OOD Performance}
We report the MSE values for the models' 5-step forecasts in Table \ref{PMSM results}. MSE values for IID data are reported on the test set of the simulated data. The baseline model performs worse across all evaluation metrics. The model with factored transition model fits best to the IID dataset and achieves the lowest MSE on OOD data. Besides, we repeatedly observe the usefulness of treating each factored representations while decoding as it performs the best (lowest). Thus, the models with better structural inductive bias can learn some form of systematic generalization and capture the complex internal dynamics as well.

\paragraph{Long Horizon Forecasting}
It is evident to question the ability of the model to forecast well up to horizons larger than those used to train them. We present Figure \ref{fig:pmsm multistep} and Figure \ref{fig: timeseries} to understand the long horizon forecasting capabilities of various models, up to 50-time steps in the future given they have been trained for 5-step predictions. 

Figure \ref{fig:pmsm multistep} shows the accumulated MSE on out-of-distribution dataset for the models in consideration. In corroboration with our previous quantitative results, we observe that the model without any structured transition models suffers the most as the prediction horizon increases (\textbf{Baseline}). This indicates the poor quality of the learned transition model and possible overfitting to the training data. Besides, all the models with separate control encoding perform well on larger horizons (Architecture 2-4 in Figure \ref{fig:TM}), with the models having additional factored latent representations performing slightly better than its counterpart on horizons between 40-50.

To understand how closely do the architectures model the ground truth future state time series, we present Figure \ref{fig: timeseries} consisting of the ground truth future state time series from the OOD dataset for the PMSM states $i_d, i_q$ along with the predictions from all the models. Figure \ref{ts_a} shows that the baseline architecture can capture the ground truth trajectory's overall pattern with some bias, which increases with time. However, Figure \ref{ts_b} indicates that the baseline model does a poor job at capturing the trajectory of $i_d$ perfectly, even for the first few forecasting steps. Providing additional prior in terms of the separate encoder for the controls helps capture the trajectories well. 

\section{Conclusion\label{conc}}

In this work, we conducted a rigorous study of systematic generalization in NN-based forecasting models for highly nonlinear dynamical systems over temporal segments of states and controls (actions). We restrict our analysis to a fully-observable environment with deterministic transition dynamics. We investigate how well different architectural inductive biases generalize to unseen combinations of the actions and find that having factored representations for states and control helps. Our finding is in corroboration with \cite{rw5} where they highlight the usefulness of having modularity in the network \cite{rw16} for language understanding. 

Our work also showcases the value that can be derived from controlled-synthetic experiments. Even though multiple architectures were generalizing well across different NARMA scenarios, careful intervention testing indicated that some architectures were capturing the underlying dependency less than the others. Throughout our study, we assumed that the number of state variables equals the number of control variables, which might not be the case in many cases. Hence, designing better architectures that incorporates structural transition model is a key future direction.

\bibliography{IEEEexample, IEEEabrv, IEEEfull}

\end{document}